%% file: SSKD-arxiv.tex
\documentclass[runningheads]{llncs}
\usepackage{graphicx}
\usepackage{comment}
\usepackage{amsmath,amssymb} 
\usepackage{color}

\usepackage[ruled,linesnumbered]{algorithm2e}
\usepackage{enumerate}
\usepackage{booktabs}
\usepackage{subfigure}

\usepackage{amssymb}

\usepackage{microtype}
\usepackage[pagebackref=true,breaklinks=true,letterpaper=true,colorlinks,bookmarks=false]{hyperref}

\newcommand{\etal}{et al.}
\newcommand{\eg}{e.g.}
\newcommand{\ie}{i.e.}

\begin{document}
\pagestyle{headings}
\mainmatter
\def\ECCVSubNumber{898}  

\title{Knowledge Distillation Meets Self-Supervision} 


\titlerunning{SSKD}
%
\author{Guodong Xu\inst{1} \and
Ziwei Liu\inst{1} \and
Xiaoxiao Li\inst{1} \and
Chen Change Loy\inst{2}}
\authorrunning{Xu et al.}
%
\institute{The Chinese University of Hong Kong \and
Nanyang Technological University\\
\email{\{xg018, zwliu, lx015\}@ie.cuhk.edu.hk}\\
\email{ccloy@ntu.edu.sg}}

\maketitle

\input{chapters/abstract.tex}

\input{chapters/intro.tex}

\input{chapters/related.tex}

\input{chapters/method.tex}

\input{chapters/exp.tex}

\section{Conclusion}

In this work, we proposed a novel framework called SSKD, the first attempt that combines self-supervision with knowledge distillation. It employs contrastive prediction as an auxiliary task to help extracting richer knowledge from teacher network. A selective transfer strategy is designed to suppress the noise in teacher knowledge. We examined our method by conducting thorough experiments on CIFAR100 and ImageNet using various architectures. Our method achieves state-of-the-art performances, demonstrating the effectiveness of our approach. Further analysis showed that our SSKD can make student more similar to teacher and work well under few-shot and noisy-label scenarios.

\clearpage
%
%
\bibliographystyle{splncs04}
\bibliography{egbib}

\newpage
\input{chapters/appendix.tex}

\end{document}

%% file: chapters/abstract.tex

\begin{abstract}

Knowledge distillation, which involves extracting the ``dark knowledge'' from a teacher network to guide the learning of a student network, has emerged as an important technique for model compression and transfer learning. Unlike previous works that exploit architecture-specific cues such as activation and attention for distillation, here we wish to explore a more general and model-agnostic approach for extracting ``richer dark knowledge'' from the pre-trained teacher model. 
We show that the seemingly different self-supervision task can serve as a simple yet powerful solution. 
For example, when performing contrastive learning between transformed entities, the noisy predictions of the teacher network reflect its intrinsic composition of semantic and pose information. By exploiting the similarity between those self-supervision signals as an auxiliary task, one can effectively transfer the hidden information from the teacher to the student. 
In this paper, we discuss practical ways to exploit those noisy self-supervision signals with selective transfer for distillation. We further show that self-supervision signals improve conventional distillation with substantial gains under few-shot and noisy-label scenarios. 
Given the richer knowledge mined from self-supervision, our knowledge distillation approach achieves state-of-the-art performance on standard benchmarks, \ie, CIFAR100 and ImageNet, under both similar-architecture and cross-architecture settings. The advantage is even more pronounced under the cross-architecture setting, where our method outperforms the state of the art CRD~\cite{crd} by an average of 2.3\% in accuracy rate on CIFAR100 across six different teacher-student pairs. The code and models are available at: \href{https://github.com/xuguodong03/SSKD}{https://github.com/xuguodong03/SSKD}.
    

\end{abstract}

%% file: chapters/intro.tex

\section{Introduction}

\begin{figure}[t]
\centering
\includegraphics[scale=0.26]{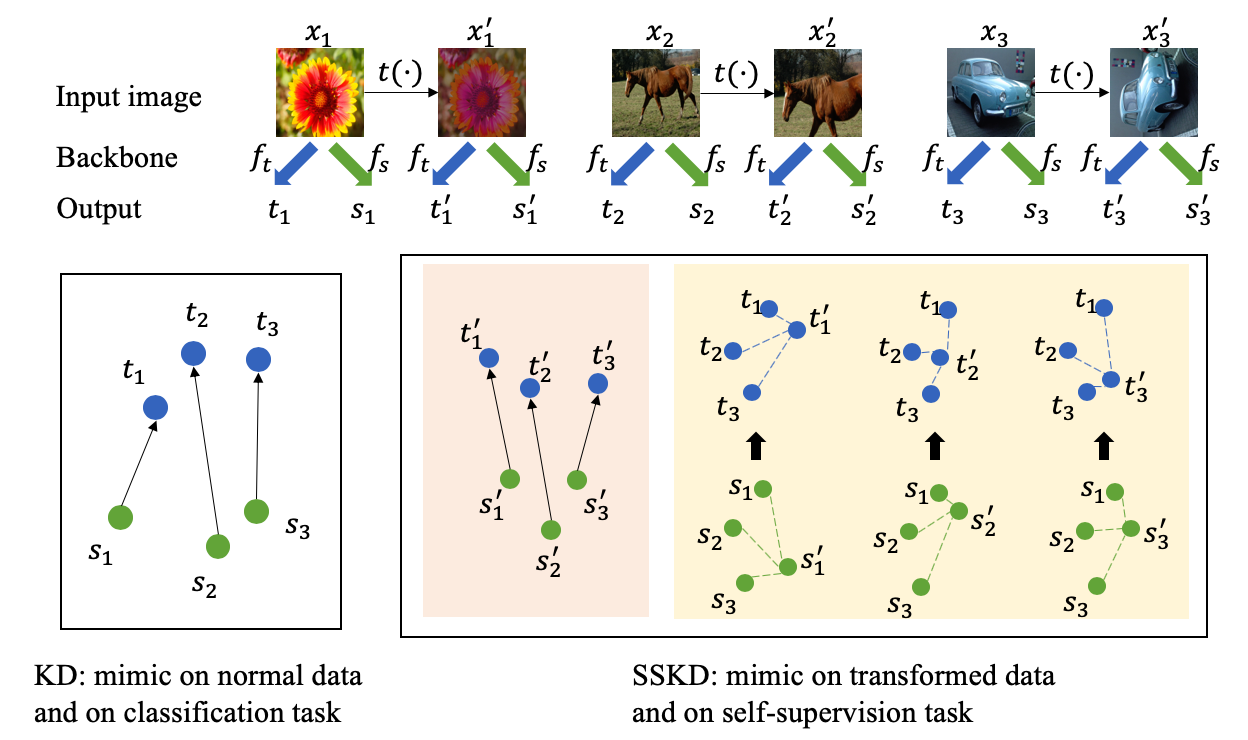}
\vspace{-8pt}
\caption{\textbf{Difference between conventional KD~\cite{KD} and SSKD}. We extend the mimicking on normal data and on a single classification task to the mimicking on transformed data and with an additional self-supervision pretext task.  The teacher's self-supervision predictions contain rich structured knowledge that can facilitate more rounded knowledge distillation on the student. In this example, contrastive learning on transformed images serves as the self-supervision pretext task. It constructs a single positive pair and several negative pairs through image transformations $t(\cdot)$, and then encourages the network to recognize the positive pair. The backbone of the teacher and student are represented as $f_t$ and $f_s$, respectively, while the corresponding output is given as $t$ and $s$ with subscript representing the index}
\label{fig:teaser}
\vspace{-0.5cm}
\end{figure}

The seminal paper by Hinton~\etal~\cite{KD} show that the knowledge from a large ensemble of models can be distilled and transferred to a student network.
Specifically, one can raise the temperature of the final softmax to produce soft targets of the teacher for guiding the training of the student. The guidance is achieved by minimizing the Kullback-Leibler (KL) divergence between teacher and student outputs.
An interesting and inspiring observation is that despite the teacher model assigns probabilities to incorrect classes, the relative probabilities of incorrect answers are exceptionally informative about generalization of the trained model.
The hidden knowledge encapsulated in these secondary probabilities is sometimes known as ``dark knowledge''.

In this work, we are fascinated on how one could extract richer ``dark knowledge'' from neural networks. Existing studies focus on what types of intermediate representations of teacher networks should student mimic.
These representations include attention map~\cite{AT}, gram matrix~\cite{FSP}, gradient~\cite{jacob}, pre-activation~\cite{AB}, and feature distribution statistics~\cite{intrakd}. 
While the intermediate representations of the network could provide more fine-grained information, a common characteristic shared by these medium of knowledge is that they are all derived from a single task (typically the original classification task).
The knowledge is highly task-specific, and hence, such knowledge may only reflect a single facet of the complete knowledge encapsulated in a cumbersome network.
To mine for richer “dark knowledge”, we need an auxiliary task apart from the original classification task, so as to extract richer information that is complementary to the classification knowledge.

In this study, we show that a seemingly different learning scheme -- self-supervised learning, when treated as an auxiliary task, can help gaining more rounded knowledge from a teacher network. 
The original goal of self-supervised learning is to learn representations with natural supervisions derived from data via a pretext task. Examples of pretext tasks include exemplar-based method~\cite{exemplar}, rotation prediction~\cite{rot}, jigsaw~\cite{jigsaw}, and contrastive learning~\cite{SimCLR,PIRL}.
To use self-supervised learning as an auxiliary task for knowledge distillation, one can apply the pretext task to a teacher by appending a lightweight auxiliary branch/module to the teacher's backbone, updating the auxiliary module with the backbone frozen, and then extract the corresponding self-supervised signals from the auxiliary module for distillation.
An example of combining a contrastive learning pretext task~\cite{SimCLR} with knowledge distillation is shown in Fig.~\ref{fig:teaser}.

The example in Fig.~\ref{fig:teaser} reveals several advantages of using self-supervised learning as an auxiliary task for knowledge distillation (we name the combination as SSKD). First, in conventional knowledge distillation, a student mimics a teacher from normal data based on a single classification task.
SSKD extends the notion to a broader extent, \ie, mimicking on transformed data and on an additional self-supervision pretext task. This enables the student to capture richer structured knowledge from the self-supervision predictions of teacher, which cannot be sufficiently captured by a single task. We show that such structured knowledge not only improves the overall distillation performance, but also regularizes the student to generalize better on few-shot and noisy-label scenarios.

Another advantage of SSKD is that it is model-agnostic.
Previous knowledge distillation methods suffer from degraded performance under cross-architecture settings, for the knowledge they transfer is very architecture-specific. For example, when transfer the feature of ResNet50~\cite{resnet} to ShuffleNet~\cite{shufflenet}, student may have trouble in mimicking due to the architecture difference.
In contrast, SSKD transfers only the last layer's outputs, hence allowing a more flexible solution space for the student model to search for intermediate representations that best suit its own architecture.

\noindent
\textbf{Contributions:} We propose a novel framework called SSKD that leverages self-supervised tasks to facilitate extraction of richer knowledge from teacher network to student network. To our knowledge, this is the first work that defines the knowledge through self-supervised tasks.
We carefully investigate the influence of different self-supervised pretext tasks and the impact of noisy self-supervised predictions to the performance of knowledge distillation. 
We show that SSKD greatly boosts the generalizability of student networks and offers significant advantages under few-shot and noisy-label scenarios.
Extensive experiments on two standard benchmarks, CIFAR100~\cite{cifar100} and ImageNet~\cite{imagenet}, demonstrate the effectiveness of SSKD over other state-of-the-art methods.

%% file: chapters/related.tex

\section{Related Work}

\noindent
{\bf Knowledge Distillation.} Knowledge distillation trains a smaller network using the supervision signals from both ground truth labels and a larger network. Hinton et al.~\cite{KD} propose to match the outputs of classifiers of two models by minimizing the KL-divergence of the category distribution. 
%
Besides the final layer logits, teacher network also distills compact feature representations from its backbone. FitNets~\cite{fitnets} proposes to mimic the intermediate feature maps of teacher network using L2 loss. AT~\cite{AT} uses attention transfer to teach student which region is the key for classification. FSP~\cite{FSP} distills the second order statistics (Gram matrix) between different layers. To alleviate information leak, FT~\cite{FT} introduces an auto-encoder in teacher network to compress features into ``factors'' and then use a translator to extract ``factors'' in student network. AB~\cite{AB} forces student to learn the binarized values of pre-activation map in teacher network.
IRG~\cite{IRG} explores whether the similarity between samples transfer more knowledge.
KDSVD~\cite{kdsvd} calls its method as self-supervised knowledge distillation. Nevertheless, the study regards the teacher's correlation maps of feature singular vectors as self-supervised labels. The label is obtained from the teacher rather than a self-supervised pretext task. Thus, their notion of self-supervised learning differ from the conventional one. Our work, to our knowledge, is the first study that investigates defining the knowledge via self-supervised pretext tasks.
CRD~\cite{crd} also combines self-supervision (SS) with knowledge distillation. The difference is the purpose of SS and how contrastive task is performed. In CRD, contrastive learning is performed across teacher and student networks to maximize the mutual information between two networks. In SSKD, contrastive task serves as a way to define knowledge. It is performed separately in two networks and then matched together through KL-divergence, which is very different from CRD.


\noindent
{\bf Self-Supervised Learning.} 
Self-supervision methods design various pretext tasks whose labels can be derived from the data itself. In the process of solving these tasks, the network learn useful representations.
Based on pretext tasks, SS methods can be grouped into several categories, including construction-based methods such as inpainting~\cite{inpainting} and colorization~\cite{color}, prediction-based methods~\cite{relative_pos,exemplar,rot,rot2,shuffle,jigsaw,jigsawpp,cmp,aet}, cluster-based methods~\cite{deepcluster,odc}, generation-based methods~\cite{bigbigan,bigan,GAN} and contrastive-based methods~\cite{SimCLR,cpcpp,PIRL,cpc,cmc}. 
Exemplar~\cite{exemplar} applies heavy transformation to each training image and treat all the images generated from the same image as a separate category. This pseudo label is used to updated the network. Jigsaw puzzle~\cite{jigsaw} splits the image into several non-overlapping patches and forces the network to recognise the shuffled order. Jigsaw++~\cite{jigsawpp} also involves SS and KD. But it utilizes knowledge transfer to boost the self-supervision performance, which solves an inverse problem of SSKD. Rotation~\cite{rot2} feeds the network with rotated images and forces it to recognise the rotation angle. To finish this task, the network has to understand the semantic information containing in the image. SimCLR~\cite{SimCLR} applies augmentation to training samples and requires the network to match original image and transformed image through contrastive loss.
Considering the excellent performance obtained by SimCLR~\cite{SimCLR}, we adopt it as our main pretext task in SSKD. However, SSKD is not limited to using only contrastive learning, many other pretext tasks~\cite{exemplar,rot2,jigsaw} can also serve the purpose. We investigate their usefulness in Sec.~\ref{sec:ablation}.

%% file: chapters/method.tex

\section{Methodology}

This section is divided into three main sections.
We start with a brief review of knowledge distillation and self-supervision in Sec.~\ref{sec:preliminaries}. For self-supervision, we discuss contrastive prediction as our desired pretext task, although SSKD is not limited to contrastive prediction.
Sec.~\ref{sec:training_scheme} specifies the training process of teacher and student model. Finally, we discuss the influence of noisy self-supervised predictions and ways to handle the noise in Sec.~\ref{sec:suppress_noise}.

\subsection{Preliminaries}
\label{sec:preliminaries}

\noindent
\textbf{Knowledge Distillation.}
Hinton \etal~\cite{KD} suggest that the soft targets predicted by a well-optimized teacher model can provide extra information, comparing to hard labels (one-hot vectors).
The relatively high probabilities assigned to wrong categories encode semantic similarities between different categories. Forcing a student to mimic teacher's prediction causes the student to learn this secondary information that cannot be expressed by hard labels alone.
To obtain the soft targets, temperature scaling is introduced in~\cite{KD} to soften the peaky softmax distribution:
\begin{equation}
p^{i}(x;\tau) = \mathrm{Softmax}(s(x);\tau) = \frac{e^{s_{i}(x)/\tau}}{\sum_k e^{s_{k}(x)/\tau}},
\label{eq1}
\end{equation}
where $x$ is the data sample, $i$ is the category index, $s_{i}(x)$ is the score logit that $x$ obtains on category $i$, and $\tau$ is the temperature.
The knowledge distillation loss $L_{kd}$ measured by KL-divergence is:
\begin{equation}
L_{kd} = -{\tau}^2\sum_{x\sim\mathcal{D}_x}\sum_{i=1}^C p_t^i(x;\tau) \log(p_s^i(x;\tau)),
\label{eq2}
\end{equation}
where $t$ and $s$ denote teacher and student models, respectively, $C$ is the total number of classes, $\mathcal{D}_x$ indicates the dataset.
The complete loss function $L$ of the student model is a linear combination of the standard cross-entropy loss $L_{ce}$ and knowledge distillation loss $L_{kd}$:
\begin{equation}
L = \lambda_1L_{ce} + \lambda_2L_{kd},
\label{eq3}
\end{equation}
where $\lambda_1$ and $\lambda_2$ are balancing weights.

\vspace{0.2cm}
\noindent
\textbf{Contrastive Prediction as Self-Supervision Task.}
Motivated by the success of contrastive prediction methods~\cite{SimCLR,cpcpp,PIRL,cpc,cmc} for self-supervised learning, we adopt contrastive prediction as the self-supervision task in our framework.
The general goal of contrastive prediction is to maximize agreement between a data point and its transformed version via a contrastive loss in latent space.

Given a mini-batch containing $N$ data points $\{x_i\}_{i=1:N}$, we apply independent transformation $t(\cdot)$ (sampled from the same distribution $\mathcal{T}$) to each data point and obtain $\{\widetilde{x}_i\}_{i=1:N}$. 
Both $x_i$ and $\widetilde{x}_i$ are fed into the teacher or student networks to extract representations $\phi_i=f(x_i), \widetilde{\phi}_i=f(\widetilde{x}_i)$. 
We follow Chen \etal~\cite{SimCLR} and add a projection head on the top of the network. 
The projection head is a 2-layer multilayer perceptron. It maps the representations into a latent space where the contrastive loss is applied, \ie, $z_i=\mathrm{MLP}(\phi_i), \widetilde{z}_i=\mathrm{MLP}(\widetilde{\phi}_i)$.

We take $(\widetilde{x}_i,x_i)$ as the positive pair and $(\widetilde{x}_i, x_k)_{k\neq i}$ as the negative pair. Given some $\widetilde{x}_i$, the contrastive prediction task is to identify the corresponding $x_i$ from the set $\{x_i\}_{i=1:N}$. 
To meet the goal, the network should maximize the similarity between positive pairs and minimize the similarity between negative pairs. 
In this work, we use a cosine similarity. If we organize the similarities between $\{\widetilde{x}_i\}$ and $\{x_i\}$ into matrix form $\mathcal{A}$, then we have:
\begin{equation}
\label{eq:similarity}
    \mathcal{A}_{i,j} = \mathrm{cosine}(\widetilde{z_i},z_j) = \frac{\mathrm{dot}(\widetilde{z_i},z_j)}{||\widetilde{z_i}||_2||z_j||_2},
\end{equation}
where $\mathcal{A}_{i,j}$ represents the similarity between $\widetilde{x}_i$ and $x_j$.
The loss of contrastive prediction is:
\begin{equation}
    L = -\sum_i \log\left(\frac{\exp(\mathrm{cosine}(\widetilde{z_i},z_i)/\tau)}{\sum_{k}\exp(\mathrm{cosine}(\widetilde{z_i},z_k)/\tau)}\right)= -\sum_i \log\left(\frac{\exp(\mathcal{A}_{i,i}/\tau)}{\sum_k \exp(\mathcal{A}_{i,k}/\tau)}\right),
    \label{eq4}
\end{equation}
where $\tau$ is another temperature parameter (can be different from $\tau$ in Eqn.~\eqref{eq1}). The loss form is similar to softmax loss and can be understood as maximizing the probability that $\widetilde{z_i}$ and $z_i$ come from a positive pair. 
In the process of matching $\{\widetilde{x}_i\}$ and $\{x_i\}$, the network learns transformation invariant representations. In SSKD, however, the main goal is not to learn representations invariant to transformations, but to exploit contrastive prediction as an auxiliary task for mining richer knowledge from the teacher model.


\subsection{Learning SSKD}
\label{sec:training_scheme}

\begin{figure}[t]
	\centering
	\includegraphics[scale=0.22]{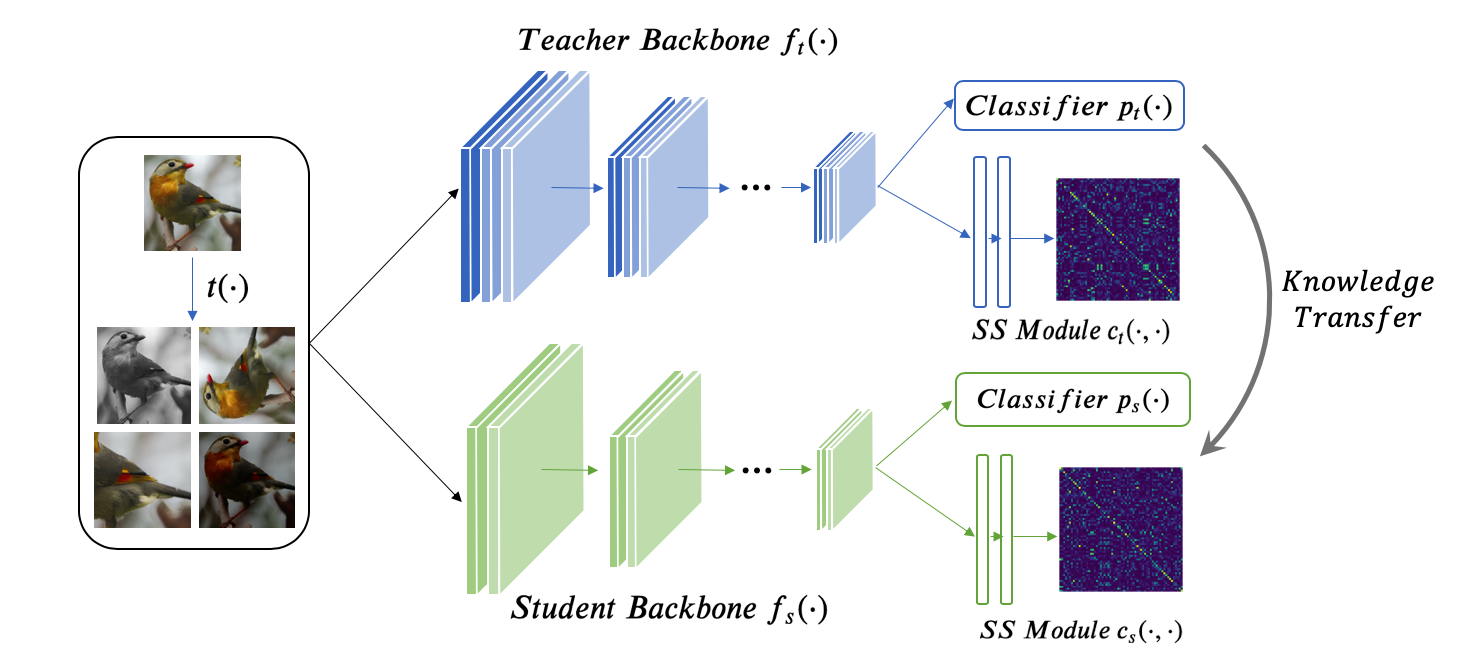}
	\vspace{-10pt}
	\caption{\textbf{Training scheme of SSKD}. Input images are transformed by designated transformations to prepare data for the self-supervision task. Teacher and student networks both contain three components, i.e., backbone $f(\cdot)$, classifier $p(\cdot)$ and SS module $c(\cdot,\cdot)$. Teacher's training are split into two stages. The first stage trains $f_t(\cdot)$ and $p_t(\cdot)$ with a classification task, and the second stage fine-tunes $c_t(\cdot,\cdot)$ with a self-supervision task. In student's training, we force the student to mimic teacher on both classification output and self-supervision output, besides the standard label loss}
	\label{fig:frm}
\end{figure}

The framework of SSKD is shown in Fig.~\ref{fig:frm}. Both teacher and student consist of three components: a backbone $f(\cdot)$ to extract representations, a classifier $p(\cdot)$ for the main task and a self-supervised (SS) module for specific self-supervision task. 
In this work, contrastive prediction is selected as the SS task, so the SS module $c_t(\cdot,\cdot)$ and $c_s(\cdot,\cdot)$ consist of a 2-layer MLP and a similarity computation module. More SS tasks will be compared in the experiments.

\vspace{0.1cm}
\noindent
\textbf{Training the Teacher Network.}
The inputs are normal data $\{x_i\}$ and transformed version $\{\widetilde{x}_i\}$. 
The transformation $t(\cdot)$ is sampled from a predefined transformation distribution $\mathcal{T}$. In this study, we select four transformations, \ie, color dropping, rotation, cropping followed by resize and color distortion, as depicted in Fig.~\ref{fig:frm}. More transformations can be included. We feed $x$ and $\widetilde{x}$ to the backbone and obtain their representations $\phi=f_t(x), \widetilde{\phi}=f_t(\widetilde{x})$. 

The training of teacher network contains two stages. 
In the first stage, the network is trained with the classification loss. Only the backbone $f_t(\cdot)$ and classifier $p_t(\cdot)$ are updated. 
Note that the classification loss is not computed on transformed data $\widetilde{x}$ because the transformation $\mathcal{T}$ is much heavier than usual data augmentation. 
Its goal is not to enlarge the training set but to make the $\widetilde{x}$ visually less similar to $x$. 
It makes the contradistinction much harder, which is beneficial to representation learning~\cite{SimCLR}. 
Forcing the network to classify $\widetilde{x}$ correctly can destroy the semantic information learned from $x$ and hurt the performance.
In the second stage, we fix $f_t(\cdot)$ and $p_t(\cdot)$, and only update parameters in SS module $c_t(\cdot,\cdot)$ using the contrastive prediction loss in Eqn.~\eqref{eq4}.

The two stages of training have distinct roles. The first stage is simply the typical training of a network for classification. The second stage, aims at adapting the SS module to use the features from the existing backbone for contrastive prediction. This allows us to extract knowledge from the SS module for distillation. It is worth pointing out that the second-stage training is highly efficient given the small MLP head, thus it is easy to prepare a teacher network for SSKD. 


\vspace{0.1cm}
\noindent
\textbf{Training the Student Network.}
After training the teacher's SS module, we apply softmax (with temperature scale $\tau$) to the teacher's similarity matrix $\mathcal{A}$ (Eqn.~\eqref{eq:similarity}) along the row dimension leading to a probability matrix $\mathcal{B}^t$, with $\mathcal{B}^t_{i,j}$ representing the probability that $\widetilde{x_i}$ and $x_j$ is a positive pair. Similar operations are applied to the student to obtain $\mathcal{B}^s$. With $\mathcal{B}^t$ and $\mathcal{B}^s$, we can compute the KL-divergence loss between the SS module's output of both teacher and student:
\begin{equation}
	\label{eq:lss}
    L_{ss} = -\tau^2 \sum_{i,j} \mathcal{B}_{i,j}^t \log(\mathcal{B}_{i,j}^s).
\end{equation}

The transformed data point $\widetilde{x}$ is the side product of contrastive prediction task. Though we do not require the student to classify them correctly, we can encourage the student's classifier output $p_s(f_s(\widetilde{x}))$ to be close to that of teacher's. The loss function is:
\begin{equation}
	\label{eq:lt}
    L_{T} = -\tau^2 \sum_{\widetilde{x}\sim\mathcal{T}(\mathcal{D}_x)}\sum_{i=1}^C p_t^i(\widetilde{x};\tau) \log(p_s^i(\widetilde{x};\tau)).
\end{equation}
The final loss for student network is the combination of aforementioned terms, \ie, cross entropy loss $L_{ce}$, $L_{kd}$ in Eqn.~\eqref{eq2}, $L_{ss}$ in Eqn.~\eqref{eq:lss}, and $L_{T}$ in Eqn.~\eqref{eq:lt}:
\begin{equation}
    L = \lambda_1 L_{ce} + \lambda_2 L_{kd} + \lambda_3 L_{ss} + \lambda_4 L_{T},
    \label{eq9}
\end{equation}
where the $\lambda_i$ is the balancing weight.


\subsection{Imperfect Self-Supervised Predictions}
\label{sec:suppress_noise}

When performing contrastive prediction, a teacher may produce inaccurate predictions, \eg, assigning $x_k$ to the $\widetilde{x_i}$, $i \neq k$. This is very likely since the backbone of the teacher is not fine-tuned together with the SS module for contrastive prediction. 
Similar to conventional knowledge distillation, those relative probabilities that the teacher assigns to incorrect answers contain rich knowledge of the teacher. 
Transferring this inaccurate but structured knowledge is the core of our SSKD.

Nevertheless, we empirically found that an extremely incorrect prediction may still mislead the learning of the student. To ameliorate negative impacts of those outliers, we adopt an heuristic approach to perform selective transfer. Specifically, we define the error level of a prediction as the ranking of the corresponding ground truth label in the classification task. Given a transformed sample $\widetilde{x}_i$ and corresponding positive pair index $i$, we sort the scores that the network assigns to each $\{x_i\}_{i=1:N}$ in a descending order. The rank of $x_i$ represents the error level of the prediction about $\widetilde{x}_i$. The rank of 1 means the prediction is completely correct. A lower rank indicates a higher degree of error.
During the training of student, we sort all the $\widetilde{x}$ in a mini-batch in an ascending order according to error levels of the teacher's prediction, and only transfer all the correct predictions and the top-$k\%$ ranked incorrect predictions. This strategy suppresses potential noise in teacher's predictions and transfer only beneficial knowledge. We show our experiments in Sec.~\ref{sec:ablation}.


%% file: chapters/exp.tex

\section{Experiments}

The experiments section consists of three parts. We first conduct ablation study to examine the effectiveness of several components of SSKD in Sec.~\ref{sec:ablation}. Comparison with state-of-the-art methods is conducted in Sec.~\ref{sec:benchmark}. In Sec.~\ref{sec:analysis}, we further show SSKD's advantages under few-shot and noisy-label scenarios. 

Evaluations are conducted on CIFAR100~\cite{cifar100} and ImageNet~\cite{imagenet} datasets, both of which are widely used as the benchmarks for knowledge distillation. CIFAR100 consists of $60,000$ $32\times32$ colour images, with $50,000$ images for training and $10,000$ images for testing. There are 100 classes, each contains 600 images. ImageNet is a large-scale classification dataset, containing $1,281,167$ images for training and $50,000$ images for testing.

\subsection{Ablation Study}
\label{sec:ablation}

\noindent
{\bf Effectiveness of Self-Supervision Auxiliary Task.}
The motivation behind SSKD is that teacher's inaccurate self-supervision output encodes rich structured knowledge of teacher network and mimicking this output can benefit student's learning. 
To examine this hypothesis, we train a student network whose only training signals come from teacher's self-supervision output, i.e., set $\lambda_1$,$\lambda_2$, $\lambda_3$ in Eqn.~(\ref{eq9}) to be 0, and observe whether student can learn good representations. 

\begin{figure}[t]
\subfigure[t-SNE visualization]{\includegraphics[scale=0.14]{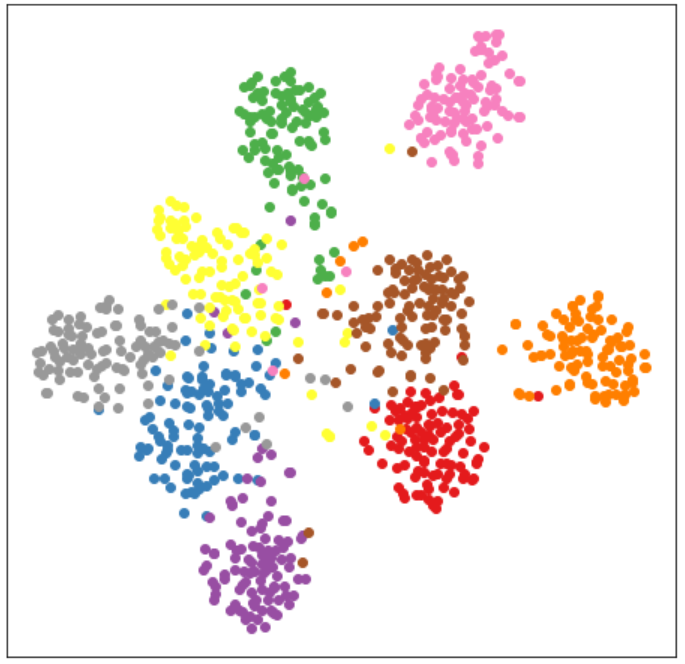}}
\subfigure[The effects of $L_T$ and $L_{ss}$]{\includegraphics[scale=0.2]{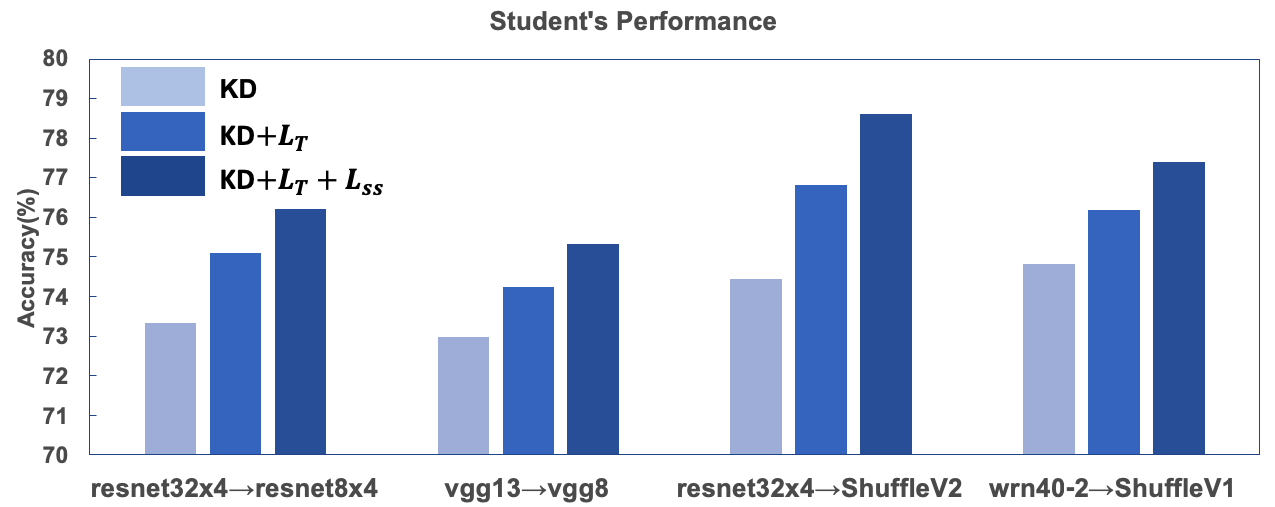}}
\vspace{-10pt}
\caption{\textbf{Effectiveness of Self-Supervision Auxiliary Task}. Mimicking the self-supervision output benefits the feature learning and final classification performance. (a) t-SNE visualization of learned features by mimicking teacher's self-supervision output. Each color represents one category. (b) The consistent improvement across all four tested teacher-student network pairs demonstrates the effectiveness of including self-supervision task as an auxiliary task}
\label{fig:ablation1}
\vspace{-0.1cm}
\end{figure}

We first demonstrate the utility by examining the student's feature distribution.
We select vgg13~\cite{vgg} and vgg8 as the teacher and student networks, respectively. The CIFAR100~\cite{cifar100} training split is selected as the training set.
After the training, we use the student backbone to extract features (before logits) of CIFAR100 test set. We randomly select 9 categories out of 100 and visualize the features with t-SNE. The results are shown in Fig.~\ref{fig:ablation1}(a). 
Though the accuracy of teacher's contrastive prediction is only around $50\%$, mimicking this inaccurate output still makes student learn highly clustered patterns, showing that teacher's self-supervision output does transfer meaningful structured knowledge.

To test the effectiveness of designed $L_T$ and $L_{ss}$, we compare three variants of SSKD with CIFAR100 on four teacher-student pairs. The three variants are: 1) conventional KD, 2) KD with additional loss $L_T$ (KD+$L_T$), 3) full SSKD (KD+$L_T$+$L_{ss}$).
The results are shown in Fig.~\ref{fig:ablation1}(b). On all four different teacher-student pairs, $L_T$ and $L_{ss}$ boost the accuracies by a large margin, showing the effectiveness of our designed components.

\noindent
{\bf Influence of Noisy Self-Supervision Predictions.}
As discussed in Sec.~\ref{sec:suppress_noise}, removing some extreme outliers are beneficial for SSKD. Some transformed samples with large error levels may play a misleading role. To examine this conjecture, we compare several students that receive different proportions of incorrect predictions from teacher. 
Specifically, we sort all the transformed $\widetilde{x}$ in a mini-batch according to their error levels in an ascending order. We transfer all the correct predictions. For incorrect predictions, we only transfer top-$k\%$ samples with the smallest error levels. A higher $k$ value indicates a higher number of predictions with larger error levels being transferred to student network.
Experiments are conducted on CIFAR100 with three teacher-student pairs. The results are shown in Table~\ref{tab:error_level}. 
%
%
The general trend shows that incorrect predictions are beneficial ($k=0$ yields the lowest accuracies). Removing extreme outliers help to give a peak performance between $k=50$ and $k=75$ across different acchitectures. 
When comparing with other methods in Sec.~\ref{sec:benchmark} and~\ref{sec:analysis}, we fix $k=75$ for all the teacher-student pairs.

\begin{table}[t]
    \centering
    \caption{\textbf{Influence of Noisy Self-Supervision Predictions to Student accuracies(\%),} when transferring the top-$k\%$ smallest error-level samples. As more samples with large error level are transferred, the performances go through a rise-and-fall process. The baseline with $k=0$ is equivalent to transferring only correct predictions}
    \vspace{-5pt}
    \begin{tabular}{l|ccccc}
    \toprule
    Teacher-Student pair & $k=0$ & $k=25$ & $k=50$ & $k=75$ & $k=100$ \\
    \midrule
    vgg13$\rightarrow$vgg8 & 74.19 & 74.36 & 74.76 & \textbf{75.01} & 74.77 \\
    resnet32$\times$4$\rightarrow$ShuffleV2 & 77.65 & 77.72 & 77.96 & \textbf{78.61} & 77.97 \\
    wrn40-2$\rightarrow$ wrn16-2 & 75.27 & 75.34 & \textbf{75.97} & 75.63 & 75.53 \\
    \bottomrule
    \end{tabular}
    \label{tab:error_level}
\end{table}

\noindent
{\bf Influence of Different Self-Supervision Tasks.}
Different pretext tasks in self-supervision would result in different qualities of extracted features. Similarly, distillation with different self-supervision tasks also lead to students with different performances. Here, we examine the influence of SS method's performance on SSKD.
We employ the commonly used linear evaluation accuracy as our metric. In particular, each method first trains a network with its own pretext task. A single layer classifier is then trained by using the representations extracted from the fixed backbone. In this way, the classification accuracies represent the quality of SS methods. In Table~\ref{tab:diff_ss}, we compare four widely used self-supervision methods: Exemplar~\cite{exemplar}, Rotation~\cite{rot2}, Jigsaw~\cite{jigsaw} and Contrastive~\cite{SimCLR}. We list the linear evaluation accuracies each method obtains on ImageNet with ResNet50~\cite{resnet} network and also student's accuracies when they are incorporated, respectively, into KD. We find that the performance of SSKD is positively correlated with the corresponding SS method.

\begin{table}[t]
    \centering
    \caption{\textbf{Influence of Different Self-Supervision Tasks}. Self-supervised (SS) performance denotes the linear evaluation accuracy on ImageNet. Student accuracies (vgg13$\rightarrow$vgg8) derived from the corresponding SS methods are positively correlated with the performance of the SS method itself. The SS performances are obtained from~\cite{SimCLR,rot2,PIRL}}
    \vspace{-5pt}
    \begin{tabular}{lcccc}
    \toprule
    SS method & Exemplar~\cite{exemplar} & Jigsaw~\cite{jigsaw} & Rotation~\cite{rot2} & Contrastive~\cite{SimCLR} \\
    \midrule
    SS performance & 31.5 & 45.7 & 48.9 & 69.3 \\
    \midrule
    
    Student performance  & 74.57 & 74.85 & 75.01 & 75.48 \\
    \bottomrule
    \end{tabular}
    \label{tab:diff_ss}
    \vspace{-0.1cm}
\end{table}

\subsection{Benchmark}
\label{sec:benchmark}


\begin{table}[t]
    \begin{minipage}{0.30\textwidth}
    \caption{\textbf{KD between Similar Architectures.} Top-1 accuracy (\%) on CIFAR100. \textbf{Bold} and \underline{underline} denote the best and the second best results, respectively. We denote by * methods that we re-run using author-provided code. SSKD obtains the best results on four out of five teacher-student pairs}
    \label{tb1}
    \end{minipage} \hspace{0.01\textwidth}
    \begin{minipage}{0.67\textwidth} 
	\begin{tabular}{cccccc}
		\toprule
		Teacher & wrn40-2 & wrn40-2 & resnet56 & resnet32$\times$4 & vgg13 \\
		Student & wrn16-2 & wrn40-1 & resnet20 & resnet8$\times$4 & vgg8 \\
		\midrule
		Teacher & 76.46 & 76.46 & 73.44 & 79.63 & 75.38 \\
		Student & 73.64 & 72.24 & 69.63 & 72.51 & 70.68 \\
		\midrule
		KD~\cite{KD} & 74.92 & 73.54 & 70.66 & 73.33 & 72.98 \\
		FitNet~\cite{fitnets} & 75.75 & 74.12 & 71.60 & 74.31 & 73.54 \\
		AT~\cite{AT} & 75.28 & 74.45 & \textbf{71.78} & 74.26 & 73.62 \\
        SP~\cite{simi} & 75.34 & 73.15 & 71.48 & 74.74 & 73.44 \\
        VID~\cite{vid} & 74.79 & 74.20 & \underline{71.71} & 74.82 & 73.96 \\
        RKD~\cite{rkd} & 75.40 & 73.87 & 71.48 & 74.47 & 73.72 \\
        PKT~\cite{pkt} & \underline{76.01} & 74.40 & 71.44 & 74.17 & 73.37 \\
        AB~\cite{AB} & 68.89 & 75.06 & 71.49 & 74.45 & 74.27 \\
        FT~\cite{FT} & 75.15 & 74.37 & 71.52 & 75.02 & 73.42 \\
		CRD*~\cite{crd} & \textbf{76.04} & \underline{75.52} & 71.68 & \underline{75.90} & \underline{74.06} \\
		\midrule
		Ours & \textbf{76.04} & \textbf{76.13} & 71.49 & \textbf{76.20} & \textbf{75.33} \\
		\bottomrule
	\end{tabular}	
    \end{minipage}
\end{table}

\noindent
\textbf{CIFAR100.}
We compare our method with representative knowledge distillation methods, including: KD~\cite{KD}, FitNet~\cite{fitnets}, AT~\cite{AT}, SP~\cite{simi}, VID~\cite{vid}, RKD~\cite{rkd}, PKT~\cite{pkt}, AB~\cite{AB}, FT~\cite{FT}, CRD~\cite{crd}.  
ResNet~\cite{resnet}, wideResNet~\cite{wrn}, vgg~\cite{vgg}, ShuffleNet~\cite{shufflenet} and MobileNet~\cite{mobilenet} are selected as the network backbones.
For all competing methods, we use the implementation of~\cite{crd}.
For a fair comparison, we combine all competing methods with conventional KD~\cite{KD} (except KD itself). And we omit ``+KD" notation in all the following tables (except for Table~\ref{tb3}) and figures for simplicity. \footnote{For experiments on CIFAR100, since we add the conventional KD with competing methods, the results are slightly better than those reported in CRD~\cite{crd}. More details on experimental setting are provided in the appendix.}

We compare performances on 11 teacher-student pairs to investigate the generalization ability of each method. Following CRD~\cite{crd}, we split these pairs into 2 groups according to whether teacher and student have similar architecture styles. The results are shown in Table~\ref{tb1} and Table~\ref{tb2}. In each table, the second partition after the header show the accuracies of the teacher's and student's performance when they are trained individually, while the third partition show the student's performance after knowledge distillation.

For teacher-student pairs with a similar architecture, SSKD performs the best in four out of five pairs (Table~\ref{tb1}).
The gap between SSKD and the best-performing competing methods is 0.52\% (averaged on five pairs). Notably, in all six teacher-student pairs with different architectures, SSKD consistently achieves the best results (Table~\ref{tb2}), surpassing the best competing methods by a large margin with an average absolute accuracy difference of 2.14\%.
Results on cross-architecture pairs clearly demonstrate that our method does not rely on architecture-specific cues. Instead, SSKD distills knowledge only from the outputs of the final layer of teacher model. Such strategy allows a larger solution space for student model to search intermediate representations that best suit its own architecture.

\begin{table*}[t]
	\centering
	\caption{\textbf{KD between Different Architectures.} Top-1 accuracy (\%) on CIFAR100. \textbf{Bold} and \underline{underline} denote the best and the second best results, respectively. We denote by * methods that we re-run using author-provided code. SSKD consistently obtains the best results on all pairs}
	\vspace{-5pt}
	\begin{tabular}{ccccccc}
		\toprule
		Teacher & vgg13 & ResNet50 & ResNet50 & resnet32$\times$4 & resnet32$\times$4 & wrn40-2 \\
		Student & MobileNetV2 & MobileNetV2 & vgg8 & ShuffleV1 & ShuffleV2 & ShuffleV1 \\
		\midrule
		Teacher & 75.38 & 79.10 & 79.10 & 79.63 & 79.63 & 76.46 \\
        Student & 65.79 & 65.79 & 70.68 & 70.77 & 73.12 & 70.77 \\
		\midrule
		KD~\cite{KD} & 67.37 & 67.35 & 73.81 & 74.07 & 74.45 & 74.83 \\
        FitNet~\cite{fitnets} & 68.58 & 68.54 & 73.84 & 74.82 & 75.11 & 75.55 \\
        AT~\cite{AT} & \underline{69.34} & 69.28 & 73.45 & 74.76 & 75.30 & 75.61 \\
        SP~\cite{simi} & 66.89 & 68.99 & 73.86 & 73.80 & 75.15 & 75.56 \\
        VID~\cite{vid} & 66.91 & 68.88 & 73.75 & 74.28 & \underline{75.78} & 75.36 \\
        RKD~\cite{rkd} & 68.50 & 68.46 & 73.73 & 74.20 & 75.74 & 75.45 \\
        PKT~\cite{pkt} & 67.89 & 68.44 & 73.53 & 74.06 & 75.18 & 75.51 \\
        AB~\cite{AB} & 68.86 & 69.32 & 74.20 & \underline{76.24} & 75.66 & \underline{76.58} \\
        FT~\cite{FT} & 69.19 & 69.01 & 73.58 & 74.31 & 74.95 & 75.18 \\
        CRD*~\cite{crd} & 68.49 & \underline{70.32} & \underline{74.42} & 75.46 & 75.72 & 75.96 \\
        \midrule
        Ours & \textbf{71.53} & \textbf{72.57} & \textbf{75.76} & \textbf{78.44} & \textbf{78.61} & \textbf{77.40} \\
		\bottomrule
	\end{tabular}
	\label{tb2}
\end{table*}

\noindent
\textbf{ImageNet.}
Limited by computation resources, we only conduct one teacher-student pair on ImageNet, i.e., ResNet34 as teacher and ResNet18 as student.
As shown in Table~\ref{tb3}, for both Top-$1$ and Top-$5$ error rates, our SSKD obtains the best performances. The results on ImageNet demonstrate the scalability of SSKD to large-scale dataset.



\noindent
\textbf{Evaluation of Learned Representations.}
Following CRD~\cite{crd}, we also investigate the qualities of student representations. A good feature extractor should generate linear separable features. Hence, we use the fixed backbone of student (trained on CIFAR100) to extract features of STL10~\cite{stl10} and TinyImageNet~\cite{tinyimagenet}, and then train a linear classifier.
We select wrn40-2 and ShuffleNetV1 as teacher and student networks, respectively.
As shown in Table~\ref{tb4}, SSKD achieves the highest accuracies on both STL10 and TinyImageNet.

\begin{table}[t]
    \small
	\centering
	\caption{\textbf{Top-1/Top-5 error (\%) on ImageNet}. \textbf{Bold} and \underline{underline} denote the best and the second best results, respectively. The competing methods include CC~\cite{cc}, SP~\cite{simi}, Online-KD~\cite{onlinekd}, KD~\cite{KD}, AT~\cite{AT}, and CRD~\cite{crd}. The results of competing methods are obtained from~\cite{crd}}
	\vspace{-5pt}
	\begin{tabular}{c|cc|ccccccc|c}
	\toprule
	& Teacher & Student & CC & SP & Online-KD & KD & AT & CRD & CRD+KD & Ours \\
	\midrule
	Top-1 & 26.70 & 30.25 & 30.04 & 29.38 & 29.45 & 29.34 & 29.30 & 28.83 & \underline{28.62} & \textbf{28.38} \\ 
	Top-5 & 8.58 & 10.93 & 10.83 & 10.20 & 10.41 & 10.12 & 10.00 & 9.87 & \underline{9.51} & \textbf{9.33}  \\
	\bottomrule
	\end{tabular}
	\label{tb3}
\end{table}

\begin{table*}[t]
    \small
	\centering
	\caption{\textbf{Linear Classification Accuracy (\%) on STL10 and TinyImageNet}. We use wrn40-2 and ShuffleNetV1 as teacher and student networks, respectively. The competing methods include KD~\cite{KD}, FitNet~\cite{fitnets}, AT~\cite{AT}, FT~\cite{FT}, and CRD~\cite{crd}}
	\vspace{-5pt}
	\begin{tabular}{cc|cc|ccccc|c}
		\hline
		\multicolumn{2}{c|}{} & Student & Teacher & KD & FitNet & AT & FT & CRD & Ours\\
		\hline
		\multicolumn{2}{l|}{CIFAR100$\rightarrow$STL10} & 71.58 & 71.01 & 73.25 & 73.77 & 73.47 & 73.56 & 74.44 & \textbf{74.74}\\
		\multicolumn{2}{l|}{CIFAR100$\rightarrow$TinyImageNet} & 32.43 & 27.74 & 32.05 & 33.28 & 33.75 & 33.69 & 34.30 & \textbf{34.54}\\
		\hline 
	\end{tabular}
	\label{tb4}
\end{table*}

\noindent
\textbf{Teacher-Student Similarity.}
SSKD can extract richer knowledge by mimicking self-supervision output and make student much more similar to teacher than other KD methods. To examine this claim, we analyze the similarity between student and teacher networks using two metrics, i.e., KL-divergence and CKA similarity~\cite{cka}. Small KL-divergence and large CKA similarity indicate that student is similar to teacher.
We use vgg13 and vgg8 as teacher and student, respectively, and use CIFAR100 as the training set. We compute the KL-divergence and CKA similarity between teacher and student on three sets, \ie, test partitions of CIFAR100, STL10~\cite{stl10} and SVHN~\cite{SVHN}.
As shown in Table~\ref{tb5}, our method achieves the smallest KL-divergence and the largest CKA similarity on CIFAR100 test set. Compared to CIFAR100, STL10 and SVHN have different distributions that have not been seen during training, therefore more difficult to mimic. However, the proposed SSKD still obtains the best results in all the metrics except KL-divergence in STL10.
From this similarity analysis, we conclude that SSKD can help student mimic teacher better and get a larger similarity to teacher network.

\begin{table}[t]
	\centering
	\small
	\caption{\textbf{Teacher-Student Similarity.} KL-divergence and CKA-similarity~\cite{cka} between student and teacher networks. \textbf{Bold} and \underline{underline} denote the best and the second best results, respectively. All the models are trained on CIFAR100 training set. $\downarrow$ ($\uparrow$) indicates the smaller (larger) the better. SSKD wins in five out of six comparisons}
	\vspace{-5pt}
	\begin{tabular}{c|cc|cc|cc}
		\hline
		Dataset & \multicolumn{2}{c|}{CIFAR100 test set} & \multicolumn{2}{c|}{STL10 test set} & \multicolumn{2}{c}{SVHN test set}\\
        \hline
		Metric & KL-div($\downarrow$) & CKA-simi($\uparrow$) & KL-div($\downarrow$) & CKA-simi($\uparrow$) & KL-div($\downarrow$) & CKA-simi($\uparrow$) \\
		\hline
		KD~\cite{KD}  & 6.91 & 0.7003 & 16.28 & 0.8234 & 15.21 & 0.6343\\
		SP~\cite{simi}  & 6.81 & 0.6816 & 16.07 & 0.8278 & 14.47 & 0.6331\\
		VID~\cite{vid}  & 6.76 & 0.6868 & 16.15 & 0.8298 & 12.60 & 0.6502\\
		FT~\cite{FT}  & 6.69 & 0.6830 & 15.95 & 0.8287 & 12.53 & \underline{0.6734}\\
		RKD~\cite{rkd}  & 6.68 & \underline{0.7010} & 16.14 & 0.8290 & 13.78 & 0.6503\\
		FitNet~\cite{fitnets}  & 6.63 & 0.6826 & 15.99 & 0.8214 & 16.34 & 0.6634\\
		AB~\cite{AB}  & 6.51 & 0.6931 & 15.34 & \underline{0.8356} & 11.13 & 0.6532\\
		AT~\cite{AT}  & 6.61 & 0.6804 & 16.32 & 0.8204 & 15.49 & 0.6505\\ 
		PKT~\cite{pkt}  & 6.73 & 0.6827 & 16.17 & 0.8232 & 14.08 & 0.6555\\
		CRD~\cite{crd}  & \underline{6.34} & 0.6878 & \textbf{14.71} & 0.8315 & \underline{10.85} & 0.6397\\
		\hline
		Ours  & \textbf{6.24} & \textbf{0.7419} & \underline{14.91} & \textbf{0.8521} & \textbf{10.58} & \textbf{0.7382}\\
		\hline 
	\end{tabular}
	\label{tb5}
\end{table}


\subsection{Further Analysis}
\label{sec:analysis}

\begin{figure}[ht]
\subfigure[Few-shot scenario]{\includegraphics[scale=0.22]{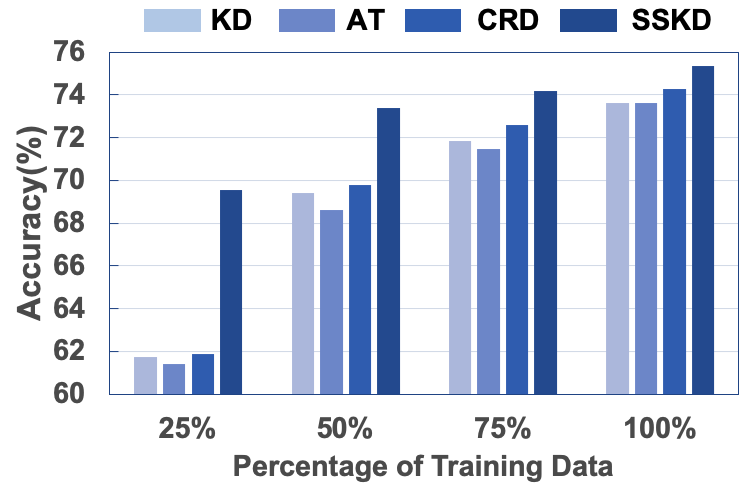}}
\subfigure[Noisy-label scenario]{\includegraphics[scale=0.22]{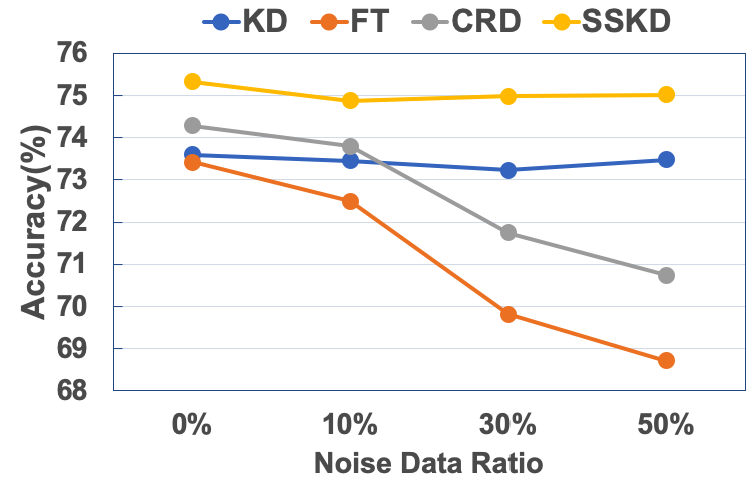}}
\vspace{-10pt}
\caption{\textbf{Accuracies on CIFAR100 test set under few-shot and noisy-label scenarios.} (a) Students are trained with subsets of CIFAR100. SSKD achieves the best results in all cases. The superiority is especially striking when only 25\% of the training data is available. (b) Students are trained with data with perturbed labels. The accuracies of FT and CRD drop dramatically as noisy labels increase, while SSKD is much more stable and maintains a high performance in all cases}
\label{fig:analysis}
\vspace{-0.1cm}
\end{figure}

\noindent
\textbf{Few-Shot Scenario.}
In a real-world setting, the number of samples available for training is often limited~\cite{liu2019large}. 
To investigate the performance of SSKD under few-shot scenarios, we conduct experiments on subsets of CIFAR100. We randomly sample images of each class to form a new training set. We train student model using newly crafted training set, while maintaining the same test set.
Vgg13 and vgg8 are chosen as teacher and student model, respectively. We compare our student's performance with KD~\cite{KD}, AT~\cite{AT} and CRD~\cite{crd}. The percentages of reserved samples are 25$\%$, 50$\%$, 75$\%$ and 100$\%$. For a fair comparison, we employ the same data for different methods.

The results are shown in Fig.~\ref{fig:analysis}(a). In all data proportions, SSKD achieves the best result. As training samples decrease, the superiority of our method becomes more apparent, \eg, $\sim7\%$ absolute improvement in accuracy compared to all competing methods when the percentage of reserved samples are 25$\%$.
Previous methods mainly focus on learning various intermediate features of teacher or exploring the relations between samples. The excessive mimicking leads to overfitting on the training set. In SSKD, the transformed images and self-supervision task endow the student model with structured knowledge that provides strong regularization, hence making it generalizes better to test set.

\noindent
\textbf{Noisy-Label Scenario.}
Our SSKD forces student to mimic teacher on both category classification task and self-supervision task. The student learns more well rounded knowledge from the teacher model than relying entirely on annotated labels. Such strategy strengthens the ability of student model to resist label noise. In this section, we investigate the performance of KD~\cite{KD}, FT~\cite{FT}, CRD~\cite{crd} and SSKD when trained with noisy label data.
We choose vgg13 and vgg8 as the teacher and student models, respectively. We assume the teacher is trained with clean data and will be shared by all students. This assumption does not affect our evaluation on robustness of different knowledge distillation methods. When training student models, we randomly perturb the labels of certain portions of training data and use the original test data for evaluation. We introduce same disturbances to all methods. Since the loss weight of cross entropy on annotated labels affects how well a model resists label noise, we use the same loss weight for all methods for a fair comparison.
We set the percentage of disturbed labels to be $0\%$, $10\%$, $30\%$ and $50\%$. Results are shown in Fig.~\ref{fig:analysis}(b). SSKD outperforms competing methods in all noise ratios. As noise data increase, the performance of FT and CRD drop dramatically. KD and SSKD are more stable.
Specifically, accuracy of SSKD only drop by a marginal $0.45\%$ when the percentage of noise data increases from $0\%$ to $50\%$, demonstrating the robustness of SSKD against noisy data labels. We attribute the robustness to the structured knowledge offered by self-supervised tasks.

%% file: chapters/appendix.tex
\section{Appendix}

\subsection{Network Architectures}

We adopt cifar-style resnet~\cite{resnet}, ResNet~\cite{resnet}, WideResNet~\cite{wrn}, MobileNet~\cite{mobilenet}, vgg~\cite{vgg} and ShuffleNet~\cite{shufflenetv2,shufflenet} as the network backbones.
For resnet, We use resnet-d to represent CIFAR-style resnet with three groups of basic blocks, each with 16, 32 and 64 channels, respectively. resnet8$\times$4 and resnet32$\times$4 indicate a $4\times$ wider network (with 64, 128 and 256 channels for each block).
For ResNet, ResNet-d represents ImageNet-style ResNet with Bottleneck blocks and more channels.
For WideResNet (wrn), wrn-d-w represents wide ResNet with depth $d$ and width factor $w$. 
For MobileNet, following ~\cite{crd}, we use a width multiplier of 0.5.
For vgg, ShuffleNetV1 and ShuffleNetV2, we adapt their architectures to CIFAR100~\cite{cifar100} dataset from their original ImageNet~\cite{imagenet} counterparts.

\subsection{Data Processing}

\noindent{\bf Data Augmentation of Normal Images.}
In CIFAR100, We pad the original image with size 4 and randomly crop it into $32\times32$. We apply horizontal flip with a probability of 0.5 and normalize three channels with the mean and standard deviation derived from the whole dataset.

In ImageNet, the images are randomly cropped, into sizes ranging from 0.08 to 1.0 of the original image size and into aspect ratios ranging from 3/4 to 4/3 of the original image aspect ratio. The cropped patch is resized to 224$\times$224 and flipped with a probability of 0.5. Finally, it is normalized by the mean and standard deviation in a channel-wise manner.

\noindent{\bf Image Transformations in Self-Supervision.}\label{sec:trans}
In SSKD, the input images contain both normal images and transformed version. We select four kinds of transformations to construct the transformation pool, namely: 1) color dropping, 2) rotation ($\pm 90^\circ,180^\circ$), 3) cropping + resizing, 4) color jitter. 

Color dropping converts an RGB image to a gray image through $L=0.229R+0.587G+0.114B$, where $L$ is the gray value. Rotation is specifically performed at three angles, \ie, $\pm 90^\circ$ and $180^\circ$, because rotation at these angles do not introduce artifacts (black regions at the image edge). Cropping + resizing first randomly crops a patch from the augmented image with size 0.08 to 1.0 and aspect ratio 3/4 to 4/3 and then resizes it to a fixed resolution (32$\times$32 in CIFAR100 and $224\times224$ in ImageNet). For color jitter, we jitter the brightness, contrast and saturation with a factor sampled from 0.6 to 1.4 and jitter the hue with a factor sampled from -0.1 to 0.1.

\subsection{Implementations of different self-supervision tasks}

In ablation study, we investigate the effects of different self-supervision tasks, \ie, Contrastive~\cite{SimCLR}, Exemplar~\cite{exemplar}, Jigsaw~\cite{jigsaw} and Rotation~\cite{rot2}. The implementation details of Contrastive are introduced in the main paper. Here we introduce the details when combining other three self-supervision tasks with knowledge distillation.

\noindent {\bf Exemplar.} Exemplar treats each instance of the dataset as a separate class. It applies heavy image transformations to the original image and forces the network to classify transformed image correctly. When combining it with KD, we adopt the same transformations as discussed in Sec \ref{sec:trans}. The SS module is a classifier with class number equalling to the dataset size. In student's training, we transfer the logits of this classifier from teacher to student.

\noindent {\bf Jigsaw.} Jigsaw splits the original image into several non-overlapping patches and permutes them. It re-organizes these patches into image format and forces the network to recognize the permutation pattern. When combining it with KD, we split each image into $2\times2=4$ patches. Thus, the SS module is a classifier with 24 ($4!=24$) classes. In student's training, we transfer the logits of this classifier.

\noindent {\bf Rotation.} Rotation first rotates the image with four angles, \ie, $0^\circ,\pm90^\circ,180^\circ$ and then forces the network to classify the rotation angle correctly. When combining it with KD, the SS module is a 4-way classifier. In student's training, we transfer the logits of this classifier from teacher to student.

\subsection{Training Hyperparameters}

\noindent{\bf Hyperparameters in Competing Methods.}
The losses of all competing methods can be uniformly written as:
\begin{equation}
    L = \alpha_1L_{ce} + \alpha_2L_{kd} + \alpha_3L_{distill},
\end{equation}
where $L_{ce}$, $L_{kd}$ and $L_{distill}$ are cross-entropy loss, conventional KD loss and specially designed loss in each method, respectively. 

In CIFAR100, we re-run all the competing methods using the implementation of CRD~\cite{crd}. We set $\alpha_1=0.1$, $\alpha_2=0.9$. For $\alpha_3$ of different methods, we use the values reported in CRD. Note that CRD sets the $\alpha_2$ in all methods to be 0, so the results of competing methods in our experiments are better than those in CRD. 
In ImageNet, we do not re-run the competing methods, but copy their results from CRD directly.

\noindent{\bf Hyperparameters in SSKD.}
In CIFAR100, we set the temperature $\tau$ in $L_{kd}$ and $L_T$ to be 4 and $\tau$ in $L_{ss}$ to be 0.5. 
For student training, we set $\lambda_1=0.1,\lambda_2=0.9,\lambda_3=2.7,\lambda_4=10.0$ in Eq 8. 
We train all the models for 240 epochs. The initial learning rate is 0.05 and is decayed by a factor of 10, respectively, at 150, 180 and 210 epochs.
We run experiments on a TITAN-X-Pascal GPU with a batch size of 64. An SGD optimizer with a $5\times10^{-4}$ weight decay and 0.9 momentum is adopted. 

\vspace{0.2cm}
In ImageNet, we use the same temperatures and loss weights as those in CIFAR100 experiments, except that $\lambda_1$ is set to 1.0. We train all the models for 100 epochs. The initial learning rate is 0.1 and is decayed by 10, respectively, at 30, 60 and 90 epochs. We train models with eight parallel GPUs with a total batch size of 256. The optimizer parameters are the same as those in CIFAR100 experiments.

\subsection{Visualization of Correlation Difference}

\begin{figure}[t]
    \centering
    \subfigure[Teacher: ResNet50, Student: MobileNetV2]
    {\includegraphics[scale=0.245]{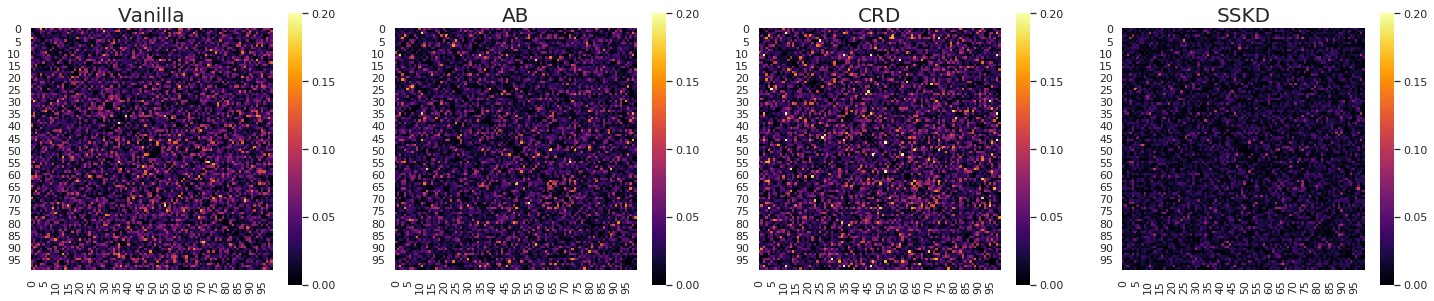}}
    \subfigure[Teacher: resnet32$\times4$, Student: ShuffleNetV2]
    {\includegraphics[scale=0.245]{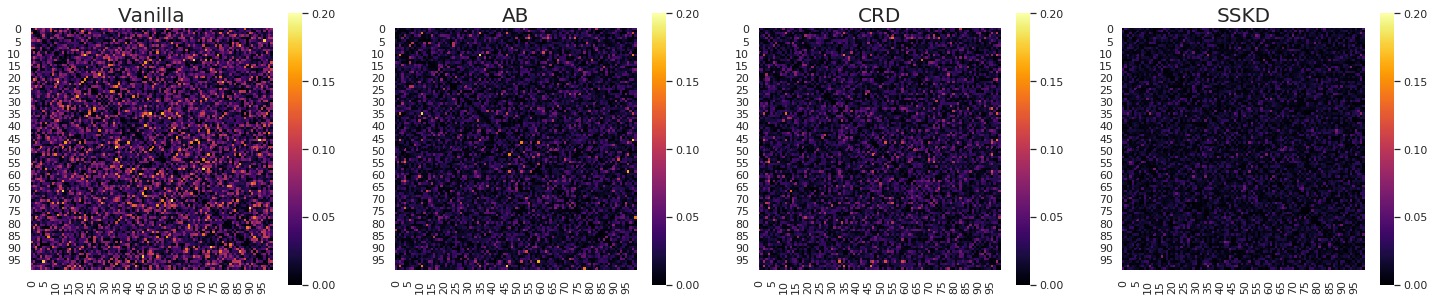}}
    \caption{The difference between correlation matrices of teacher's and student's classifier weights (of CIFAR100). The correlation matrices are computed using normalized weights. We select two teacher-student pairs and four methods: Vanilla student, AB~\cite{AB}, CRD~\cite{crd} and our SSKD. SSKD achieves significant matching between student's and teacher's correlations, indicating that it captures the teacher's correlation structures best}
    \label{fig:1}
\end{figure}

We compute the difference (L1 error) between the correlation matrices of the teacher’s and student’s classifier weights and visualize the difference through heatmap (Fig.~\ref{fig:1}). We select four methods: vanilla student without distillation and students trained by AB~\cite{AB}, CRD~\cite{crd} and our SSKD. It is clear that SSKD obtains the smallest difference on both two teacher-student pairs, indicating that SSKD captures the teacher's correlation structure best.